%% file: main.tex
\documentclass[conference]{IEEEtran}
\usepackage{spconf,amsmath,graphicx}

\renewcommand{\thetable}{\arabic{table}}

\usepackage{geometry}
\usepackage{enumitem}
\setlist{nosep, leftmargin=14pt}

\usepackage{mwe} %

\usepackage{xspace}

\makeatletter
\DeclareRobustCommand\onedot{\futurelet\@let@token\@onedot}
\def\@onedot{\ifx\@let@token.\else.\null\fi\xspace}

\def\ie{\emph{i.e}\onedot} 
 
\def\etc{\emph{etc}\onedot} 
 
\def\etal{\emph{et al}\onedot}
\makeatother

\usepackage[noadjust]{cite} %

\usepackage{amssymb}
\usepackage{amsfonts}
\usepackage{caption}
\usepackage{subcaption}

\usepackage{multirow}
\usepackage{makecell}
\usepackage{utfsym}
\newcommand{\crossmark}{\scalebox{0.75}{\usym{2613}}}

\usepackage[capitalize]{cleveref}
\crefname{section}{Sec.}{Secs.}
\Crefname{section}{Section}{Sections}
\Crefname{table}{Table}{Tables}
\crefname{table}{Tab.}{Tabs.}

\title{DentalX: Context-aware Dental Disease Detection with Radiographs}
\name{Zhi Qin Tan$^{\star}$ \qquad Xiatian Zhu$^{\dagger}$ \qquad Owen Addison$^{\star}$ \qquad Yunpeng Li$^{\star}$}

\address{$^{\star}$ Centre for Oral, Clinical \& Translational Sciences, King’s College London, United Kingdom \\
    $^{\dagger}$ Surrey Institute for People-Centred AI, University of Surrey, United Kingdom} 
\begin{document}
\maketitle
\input{sec/0_abstract}    
\input{sec/1_intro}

\input{sec/2_method}
\input{sec/3_experiments}
\input{sec/4_conclusion}

\section{Compliance with ethical standards}
\label{sec:ethics}
This study was performed in line with the principles of the Declaration of Helsinki. Approval was granted by the Ethics Committee of the University of Surrey (Date 13.10.2023 / FEPS 22-23 035 EGA).

\section{Acknowledgments}
\label{sec:acknowledgments}
This work was supported by the National Institute for Health and Care Research i4i Programme (NIHR204566).

\let\OLDthebibliography\thebibliography
\renewcommand\thebibliography[1]{
  \OLDthebibliography{#1}
  \setlength{\parskip}{0pt}
  \setlength{\itemsep}{0pt plus 0.3ex}
}

\normalsize{
    \bibliographystyle{IEEEbib}
    \bibliography{refs}
}
\end{document}

%% file: sec/0_abstract.tex
\begin{abstract}
Diagnosing dental diseases from radiographs is time-consum\-ing and challenging due to the subtle nature of diagnostic evidence. Existing methods, which rely on object detection models designed for natural images with more distinct target patterns, struggle to detect dental diseases that present with far less visual support.
To address this challenge, we propose {\bf DentalX}, a novel context-aware dental disease detection approach that leverages oral structure information to mitigate the visual ambiguity inherent in radiographs. Specifically, we introduce a structural context extraction module that learns an auxiliary task: semantic segmentation of dental anatomy. The module extracts meaningful structural context and integrates it into the primary disease detection task to enhance the detection of subtle dental diseases.
Extensive experiments on a dedicated benchmark demonstrate that DentalX significantly outperforms prior methods in both tasks.
This mutual benefit arises naturally during model optimization, as the correlation between the two tasks is effectively captured. 
Our code is available at https://github.com/zhiqin1998/DentYOLOX.

\end{abstract}

\begin{keywords}
Context-aware learning, Dental disease detection, Dental anatomy segmentation
\end{keywords}

%% file: sec/1_intro.tex
\section{Introduction}
Dental radiographs, such as bitewing and periapical radiographs, are essential in diagnosing dental diseases such as tooth caries, periodontal bone loss, and periapical lesions. However, diagnosing dental diseases is a tedious and challenging task as high variability between observers exists even among experienced experts \cite{caries_interobs_var}, due to 
the subtlety between the various severity of the diseases. On the other hand, deep learning has shown remarkable potential in the detection and diagnosis of dental diseases from dental radiographs \cite{CARVALHO2024review}. Therefore, the demand for deep learning techniques as a useful aid in improving dentists' diagnostic performance is growing rapidly in clinical settings.

As convolutional neural networks (CNNs) have been widely adopted in various medical image diagnostic tasks, many studies have recently been made in the dental disease detection task. Existing works include adopting Faster R-CNN \cite{fasterrcnn2015} for the detection of marginal bone loss \cite{Liu2022marginalbl} and caries \cite{zhu2022caries} on periapical radiographs. The YOLO \cite{yolo} model was employed to detect caries in bitewing radiographs \cite{Bayraktar2022caries} while Chen \etal~\cite{Chen2021caries_bl} investigated the effectiveness of Faster R-CNN in detecting three types of caries, periodontal bone loss and periapical lesion. Besides, Cha \etal~\cite{cha2021bl} proposed to detect the keypoints of tooth landmarks, then the severity of periodontal bone loss is computed based on the ratio of bone level and tooth length. Moreover, several studies have investigated the use of Faster R-CNN and DetectNet for the detection of lesions \cite{Lee2022caries_pano,ba2023perilesion_pano} and periodontal bone loss \cite{Jiang2022bl_pano} on panoramic radiographs.
Designed for natural images with clear inter-class object distinction, existing object detectors would suffer when learning to detect highly fine-grained and subtle dental disease classes in radiographic images.

To address this fundamental limitation, we investigate the integration of contextual knowledge to optimize the disease detector by utilizing dental structural information. This approach aims to reduce the inherent visual ambiguity in radiographic images. 
This is aligned with dentists' assessment procedures, which often first or simultaneously assess oral anatomy while diagnosing diseases from dental radiographs.
Our motivation also stems from the understanding that certain dental diseases can only manifest on specific dental structures.
For example, grades 1 and 2 dental caries are confined to the tooth enamel, while grades 3 to 5 caries extend into the tooth dentin \cite{caries}. Additionally, periodontal bone loss occurs around the cemento-enamel junction of the tooth \cite{perio_bl}. Integrating this structural knowledge with visual observations would enable us to explore their complementary effects, as they offer distinct types of information for decision-making.

\begin{figure*}[tb]
    \centering
    \includegraphics[width=.81\linewidth]{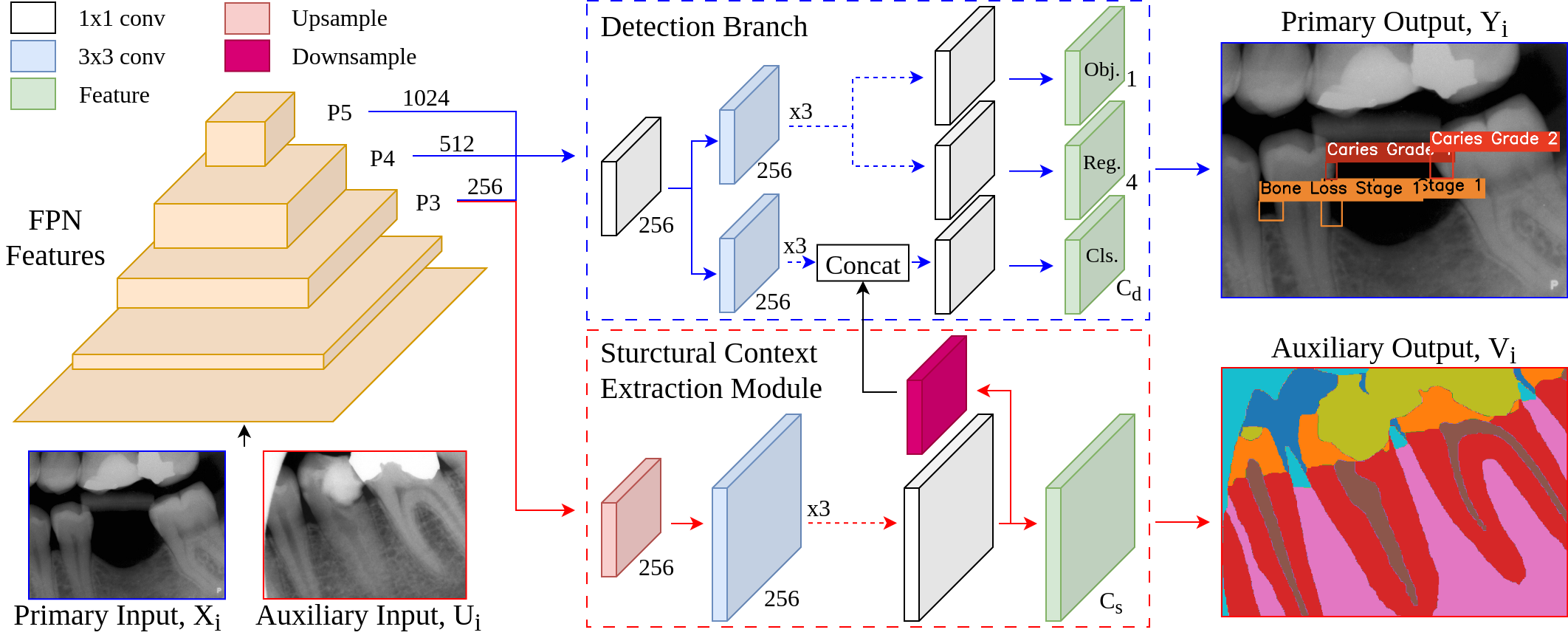}
    \caption{Overview of the proposed DentalX that jointly learns both dental disease detection and the auxiliary anatomy segmentation tasks from partially annotated data.
    The structural context extraction module extracts useful structural information from the auxiliary task to enhance disease detection.
    The numbers below each layer represent the number of output channels. 
    }
    \label{fig:overall_arch}
\end{figure*}

The contributions of this study are as follows:
\begin{enumerate}
\item We propose a context-aware dental disease detection approach, called DentalX, which leverages oral structure information to enhance the detection of subtle dental diseases in radiographic images.
\item We implement DentalX by designing a dental disease detection architecture that incorporates structural context through a jointly learned auxiliary dental anatomy segmentation task. Specifically, we introduce a structural context extraction module that creates a unidirectional connection from the auxiliary task to the detection head to integrate anatomy information in disease detection.
\item Extensive experiments demonstrate that DentalX outperforms previous alternatives in both dental disease detection and dental anatomy segmentation, highlighting the mutual benefits of jointly learning these tasks and the model's effectiveness in capturing their correlation.
\end{enumerate}

%% file: sec/2_method.tex
\section{Method}

\subsection{Problem Formulation}
We assume a dental disease detection dataset comprising a set of $N_d$ radiographs denoted by $\mathbf{S}_d = \{(X_i, Y_i)\}_{i=1}^{N_d}$ where $X_i$ is the $i$-th radiograph and $Y_i$ contains the corresponding set of ground truth bounding boxes and classes of all diseases present in $X_i$. 
Consider another distinct dental anatomy segmentation dataset containing a set of $N_s$ radiographs denoted by $\mathbf{S}_s = \{(U_i, V_i)\}_{i=1}^{N_s}$. Here, $U_i$ is the $i$-th radiograph while $V_i$ is its associated ground truth dental anatomy segmentation mask with the same size as $U_i$. We aim to jointly learn both dental disease detection and anatomy segmentation tasks.

\subsection{Model Architecture}
\cref{fig:overall_arch} visualizes the overall model architecture of DentalX. 
The model comprises a feature pyramid network (FPN) backbone, an object detection branch, and a structural context extraction (SCE) module that learns the auxiliary anatomy segmentation task while extracting structural context information to supplement the detection branch.

\noindent\textbf{FPN Backbone:}
The FPN backbone leverages a pyramidal feature hierarchy, building multiple feature maps with rich semantic information. It consists of a bottom-up pathway 
that produces increasingly stronger but smaller feature maps and a top-down pathway that generates higher resolution feature maps through upsampling convolutions. 
FPN outputs feature maps at three decreasing scales, P3, P4, and P5, containing 256, 512, and 1024 output channels, respectively.

\noindent\textbf{Disease Detection Branch:}
The detection branch takes P3, P4, and P5 feature maps as inputs to produce object predictions on various scales. They are first reduced to 256 channels with a $1\times1$ convolution block, then passed through three $3\times3$ convolution blocks followed by a $1\times1$ convolution layer to predict the objectness score, bounding box, and class probabilities. We use a decoupled head that separates the classification head from the box regression and objectness heads to promote model convergence and improve the detection performance \cite{yolox2021}. Lastly, the detection branch receives dental structure information from the SCE module as a feature map and concatenates it with the penultimate feature map of the classification head to predict the disease class probabilities.

\noindent\textbf{Structural Context Extraction Module:}
We introduce structural context extraction (SCE) module in DentalX, which enables the segmentation of dental anatomy as an auxiliary task. The FPN feature map from P3 is used as input in SCE because it captures the richest semantic information in the top-down pathway (\ie from P4 and P5) and has the highest resolution compared to P4 and P5, a critical aspect for pixel-accurate semantic segmentation. 
The P3 feature map is first upsampled to double its spatial size, followed by three $3\times3$ and a $1\times1$ convolutional layers. The last convolution layer produces an auxiliary output with $C_s$ channels, where $C_s$ is the number of classes in the segmentation task, which include enamel, dentin, root dentin, pulp, bone, and implants. 

Next, we implicitly integrate the domain knowledge of diseases manifesting only on particular dental structures, by extracting the final feature map to the detection branch. This extraction step is lightweight and consists of only several downsampling operations to generate feature maps with appropriate sizes that match the multi-scale feature maps in the detection branch. The resized feature maps are concatenated with the penultimate output feature maps of the classification head, before passing to its final layer. This allows the detection head to utilize the semantic features of dental anatomies, taking into account the oral structure present at each spatial location when predicting the class of the dental disease. 

\subsection{Joint Learning with Partially Annotated Data}
Since datasets for different tasks are usually collected and labeled separately, there is no overlap between the training images for multiple tasks (\ie each image consists of only ground truth labels for a single task). Thus, we design the model training procedure to accommodate joint learning of both tasks with partially annotated data.
Specifically, during model training, each mini-batch contains an equal proportion of training samples from the two tasks. We compute the loss value of each training sample as $\mathcal{L} = {\mathcal{L}_d}^{i} + {\mathcal{L}_s}^{i}$.
If the $i$-th sample belongs to the detection dataset, $\mathbf{S}_d$, then ${\mathcal{L}_s}^{i} = 0$ and ${\mathcal{L}_d}^i$ is the loss value of the $i$-th sample for the detection task computed with the YOLO loss function \cite{yolox2021} which includes the regression loss for the box regression head, and the cross-entropy loss for the objectness and classification heads. Conversely, if the $i$-th sample belongs to the segmentation dataset, $\mathbf{S}_s$, then ${\mathcal{L}_d}^i = 0$ and ${\mathcal{L}_s}^i$ represents the loss value of the $i$-th sample for the auxiliary segmentation task consisting of the pixel-wise cross-entropy loss and the IoU loss.

%% file: sec/3_experiments.tex
\section{Experiments}

\noindent\textbf{Dental Disease Detection Dataset:} Our internal dental disease detection dataset consists of 10,121 anonymized periapical and bitewing radiographs originating from three dental practices. 32 dentists of various backgrounds and levels of experience annotated the dataset with 20 types of dental disease such as five grades of caries, four stages of bone loss, calculus, periapical lesion, internal/external resorption, \etc, alongside their bounding boxes. We randomly assigned five dentists to annotate each radiograph. The reference ground truth is generated via a majority voting procedure.
Lastly, 600 images were randomly selected as the test dataset with equal proportions from the three dental practices.

\noindent\textbf{Dental Anatomy Segmentation Dataset:} 
The dental anatomy segmentation dataset is a combination of periapical and bitewing radiograph images from several public datasets \cite{Rad2016dentalxraydatabase,rashi2022ahybr,fatima2022dentaldataset,kunt2023automaticdataset,azleen2024tooth}. The images from these datasets are merged into a total of 1,556 periapical radiographs and 4,388 bitewing radiographs, and their ground truth masks were labeled with six classes of dental anatomy (\ie enamel, dentin, root dentin, pulp, bone, and implants). 200 images of each type of dental radiograph are randomly selected as the test dataset.%

\noindent\textbf{Implementation Details:}
The proposed model is implemented in PyTorch 1.7 with an RTX 4090 GPU. The stochastic gradient descent optimizer is used with an initial learning rate of 0.001 and a cosine annealing policy. The model is trained for 200 epochs with a batch size of 16.

\noindent\textbf{Evaluation Metrics:}
We use the mean average precision (AP) \cite{cocodataset} to evaluate the detection performance across intersection over union (IoU) thresholds of 0.5, 0.75, and the average of $\{0.5, 0.55, ..., 0.95\}$. For segmentation metrics, the mean IoU, Dice score, and pixel accuracy are reported.

\subsection{Results and Discussion}
\label{subsec:results}
As previous works on multi-task learning of dental disease detection and anatomy segmentation focus on different dental image modalities \cite{pano-multi,cbct-multi}, we compare DentalX against previous baselines on both tasks separately. Faster R-CNN \cite{fasterrcnn2015}, YOLOX \cite{yolox2021}, and DINOv3 \cite{dinov3} are selected as the baseline detectors, while UNet \cite{unet}, SETR \cite{setr}, and PIDNet \cite{pidnet} are chosen for the segmentation task. %
All models follow their original hyperparameters and are trained with the same GPU budget for fair comparison.

\begin{table}[tb]
    \caption{Dental disease detection results.%
    }
    \label{tab:result_dd}
    \centering \setlength{\tabcolsep}{4pt}%
    \resizebox{\linewidth}{!}{
    \begin{tabular}{l|ccccc}
         \hline
         Method & AP$^{.5}$ & AP$^{.75}$ & AP$^{.5:.95}$ & Params & {Inference FPS} \\
         \hline \hline
         Faster R-CNN \cite{fasterrcnn2015} & 34.2 & 8.1 & 13.9 & 42M & 17 \\
         YOLOX \cite{yolox2021} & 40.7 & 13.0 & 18.1 & 54.2M & \textbf{191} \\
         DINOv3 \cite{dinov3} & 30.3 & 6.1 & 12.8 & 7B & 14 \\
         \hline
         DentalX (ours) & \textbf{45.9} & \textbf{17.1} & \textbf{21.0} & 54.5M & 180 \\
         \hline
    \end{tabular}
    }
\end{table}

\begin{table}[tb]
    \caption{Dental anatomy segmentation results. %
    }
    \label{tab:result_as}
    \centering \setlength{\tabcolsep}{4pt}%
    \resizebox{0.6\linewidth}{!}{
    \begin{tabular}{l|ccc}
         \hline
         Method & mIoU & mDice & mAcc \\
         \hline \hline
         UNet \cite{unet} & 81.2 & 89.5 & 90.3 \\
         SETR \cite{setr} & 84.7 & 91.5 & 91.1  \\
         PIDNet \cite{pidnet} & 82.8 & 90.5 & 92.2 \\
         \hline
         DentalX (ours) & \textbf{86.3} & \textbf{92.5} & \textbf{94.2} \\
         \hline
    \end{tabular}
    }
\end{table}

\begin{figure*}[tb]
    \begin{subfigure}{0.16\textwidth}
        \centering \includegraphics[width=\linewidth,trim={2cm 18cm 0cm 0cm},clip]{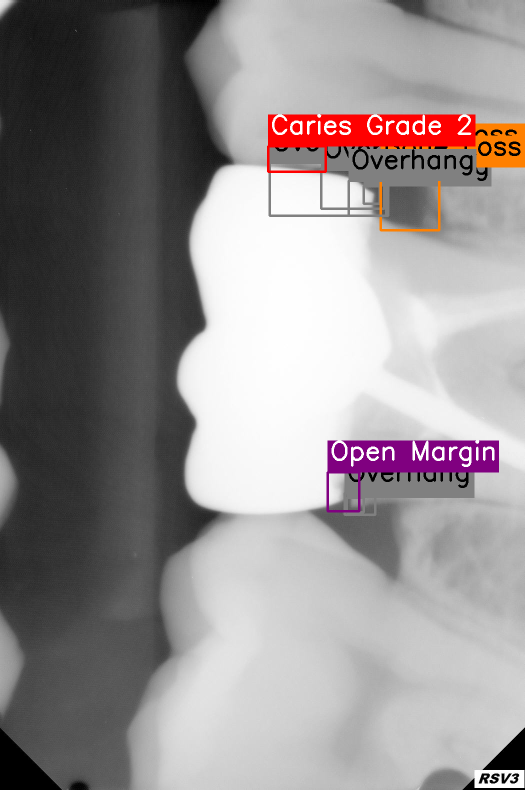}
    \end{subfigure}\hfill
    \begin{subfigure}{0.16\textwidth}
        \centering \includegraphics[width=\linewidth,trim={2cm 18cm 0cm 0cm},clip]{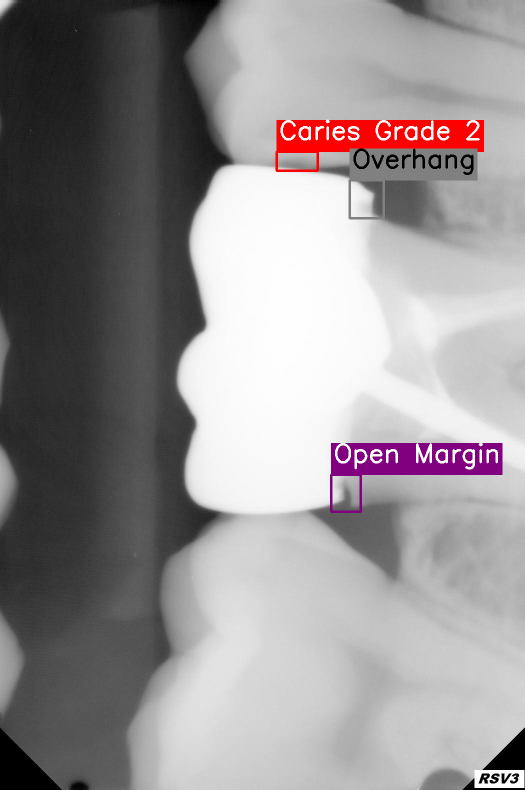}
    \end{subfigure}\hfill
    \begin{subfigure}{0.16\textwidth}
        \centering \includegraphics[width=\linewidth,trim={2cm 18cm 0cm 0cm},clip]{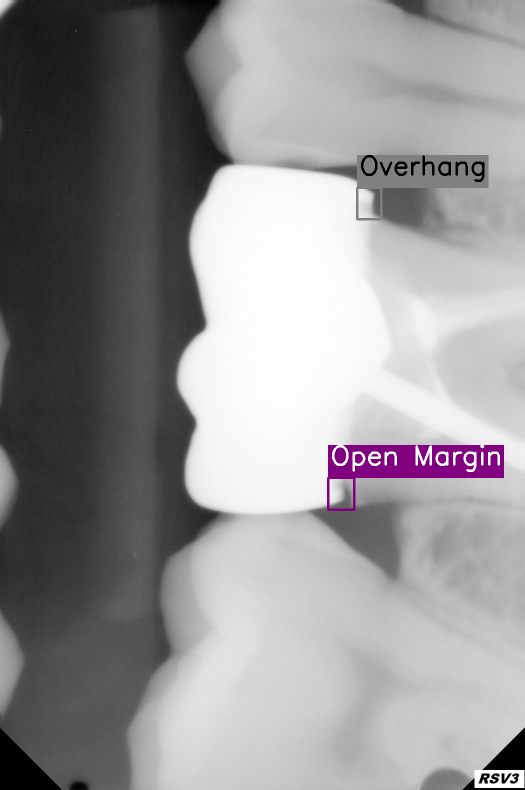}
    \end{subfigure}\hfill
    \begin{subfigure}{0.16\textwidth}
        \centering \includegraphics[width=\linewidth,trim={2cm 18cm 0cm 0cm},clip]{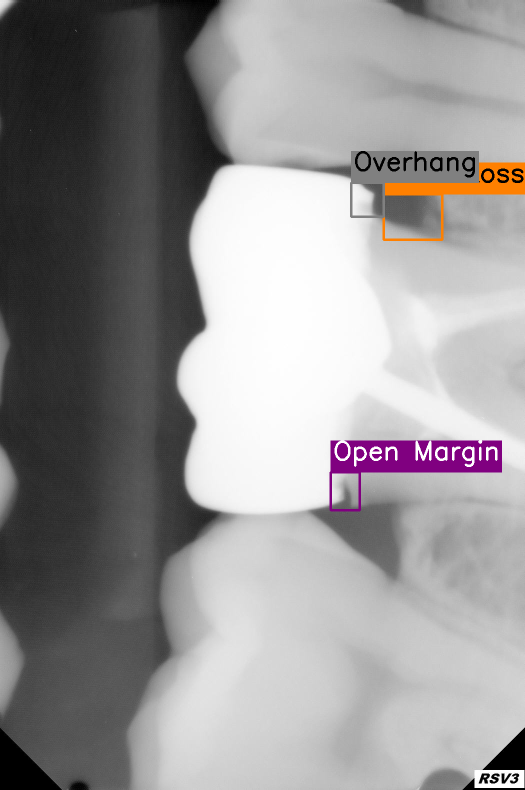}
    \end{subfigure}\hfill
    \begin{subfigure}{0.16\textwidth}
        \centering \includegraphics[width=\linewidth,trim={2cm 18cm 0cm 0cm},clip]{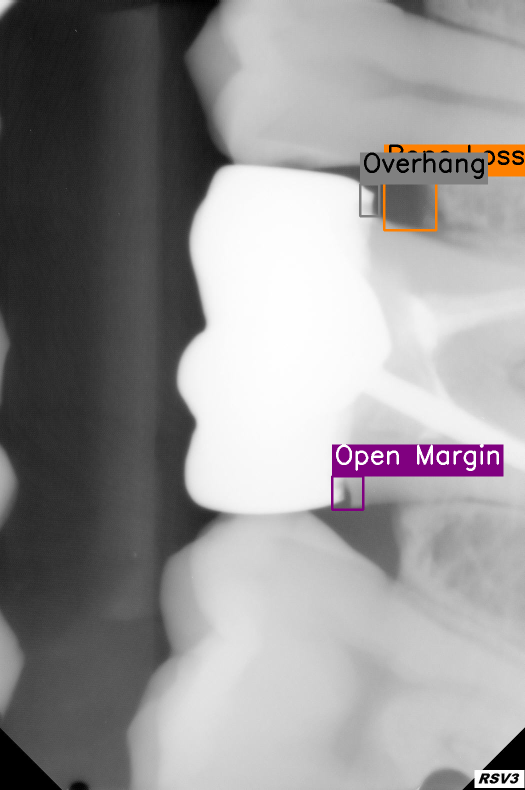}
    \end{subfigure}
    
    \vspace{0.05em}
    
    \begin{subfigure}{0.16\textwidth}
        \centering \includegraphics[width=\linewidth,trim={5cm 2cm 0cm 0.5cm},clip]{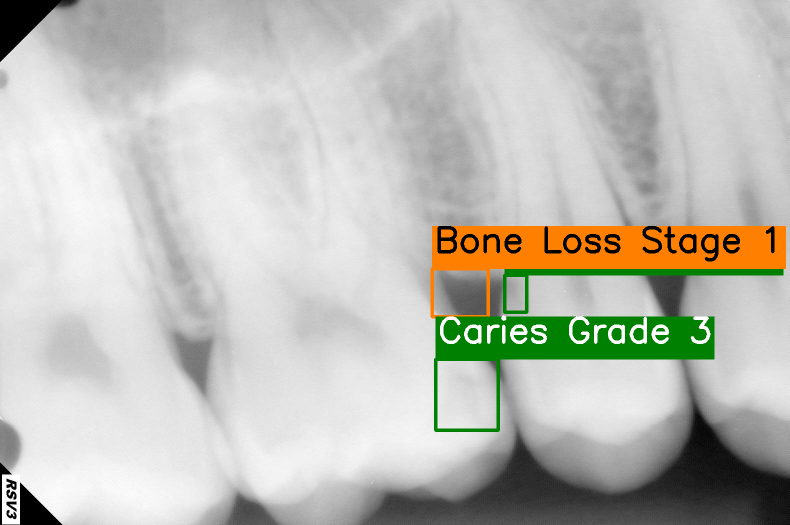}
        \caption{Faster R-CNN}
    \end{subfigure}\hfill
    \begin{subfigure}{0.16\textwidth}
        \centering \includegraphics[width=\linewidth,trim={5cm 2cm 0cm 0.5cm},clip]{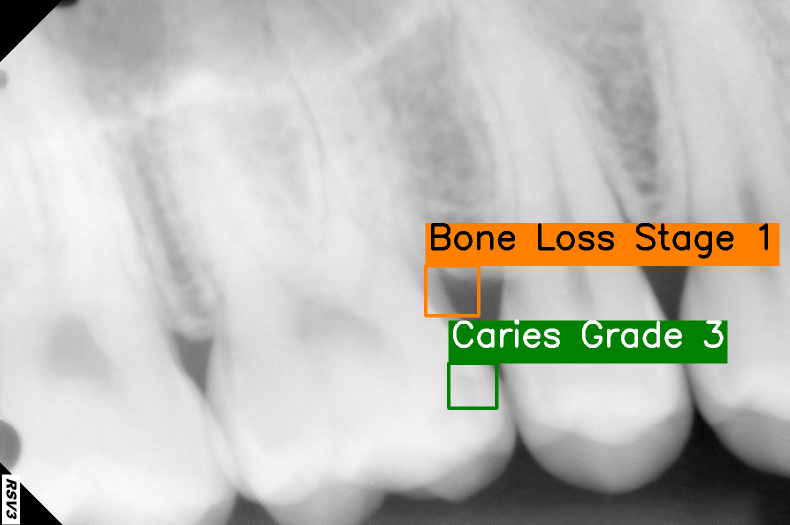}
        \caption{YOLOX}
    \end{subfigure}\hfill
    \begin{subfigure}{0.16\textwidth}
        \centering \includegraphics[width=\linewidth,trim={5cm 2cm 0cm 0.5cm},clip]{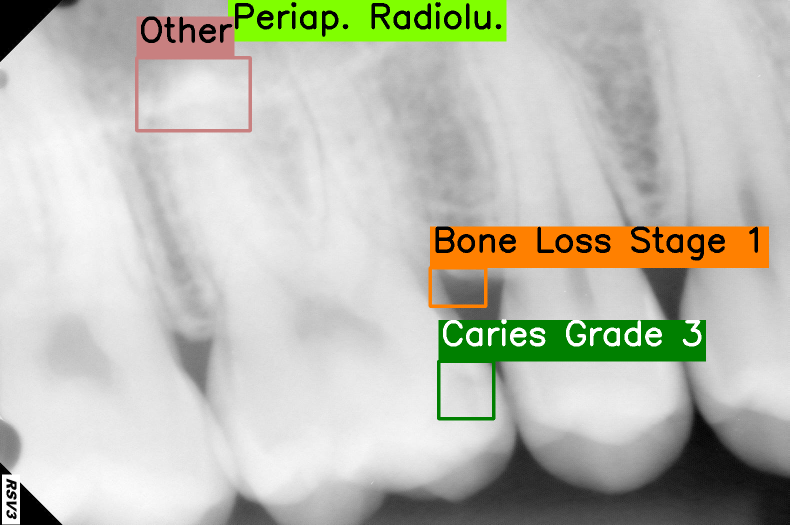}
        \caption{DINOv3}
    \end{subfigure}\hfill
    \begin{subfigure}{0.16\textwidth}
        \centering \includegraphics[width=\linewidth,trim={5cm 2cm 0cm 0.5cm},clip]{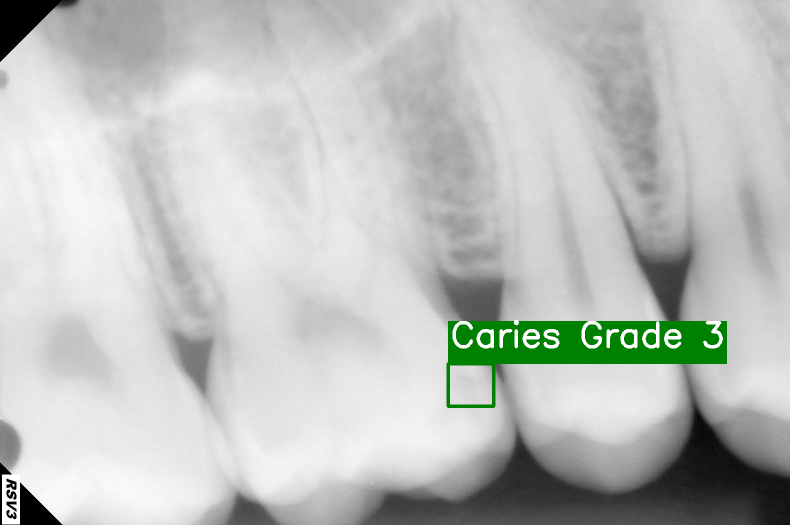}
        \caption{DentalX}
    \end{subfigure}\hfill
    \begin{subfigure}{0.16\textwidth}
        \centering \includegraphics[width=\linewidth,trim={5cm 2cm 0cm 0.5cm},clip]{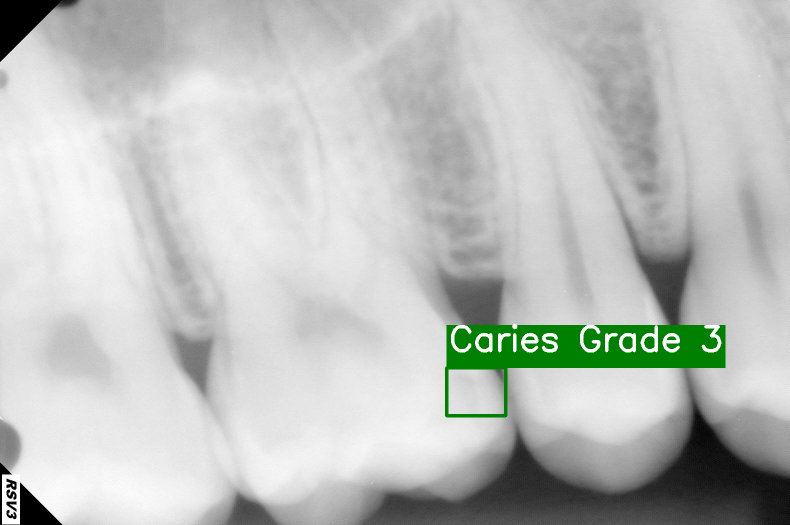}
        \caption{GT}
    \end{subfigure}
    
    \caption{(a) - (d) Dental disease detection results.
    (e) The reference ground truth (GT). 
    }
    \label{fig:det_qualitative}
\end{figure*}

\begin{figure*}[tb]
    \begin{subfigure}{\textwidth}
        \centering \includegraphics[width=.8\linewidth]{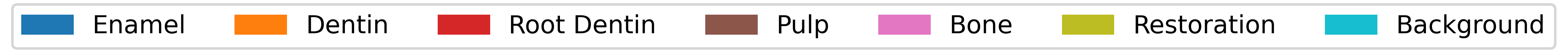}
    \end{subfigure}
    \begin{subfigure}{0.14\textwidth}
        \centering \includegraphics[width=\linewidth,trim={0.5cm 1.25cm 0.1cm 0},clip]{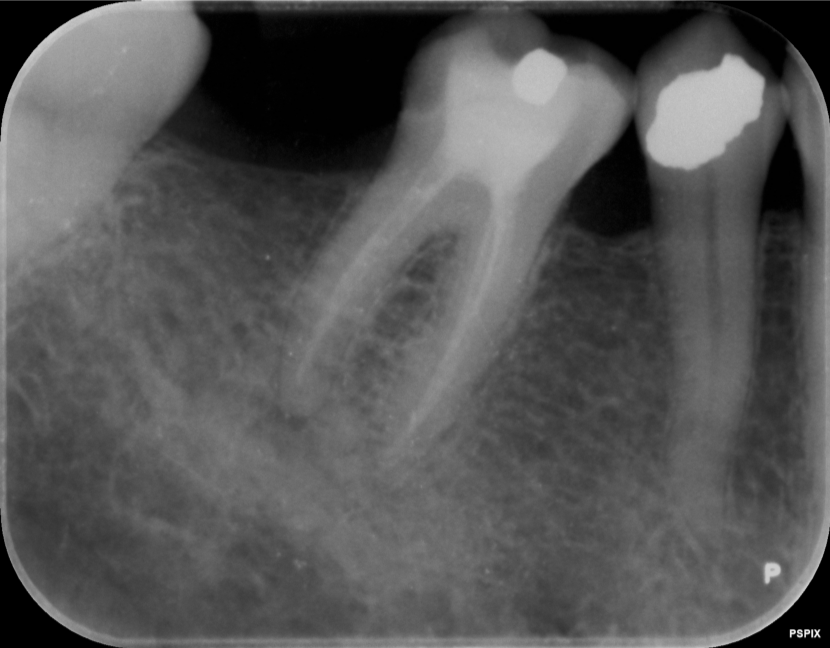}
    \end{subfigure}\hfill
    \begin{subfigure}{0.14\textwidth}
        \centering \includegraphics[width=\linewidth,trim={0.5cm 1.25cm 0.1cm 0},clip]{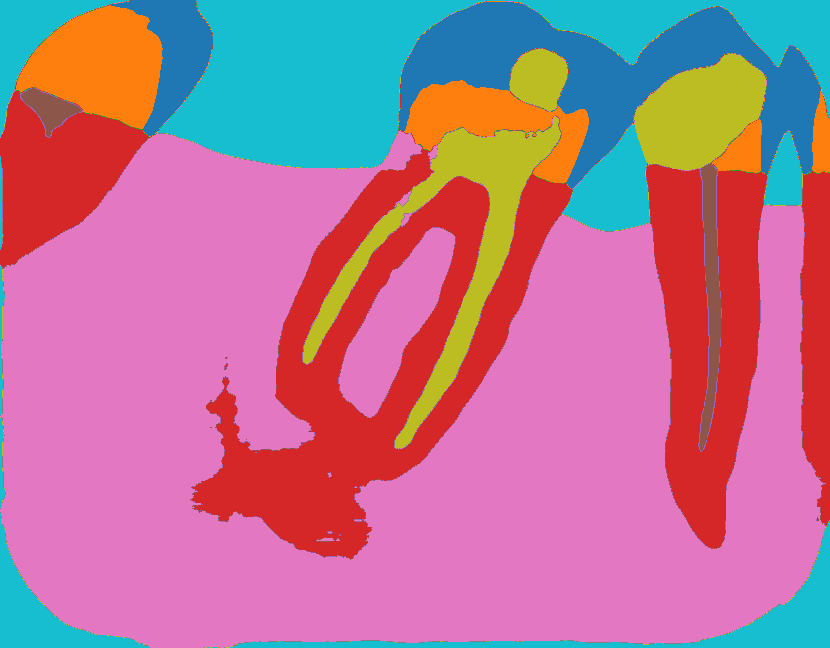}
    \end{subfigure}\hfill
    \begin{subfigure}{0.14\textwidth}
        \centering \includegraphics[width=\linewidth,trim={0.5cm 1.25cm 0.1cm 0},clip]{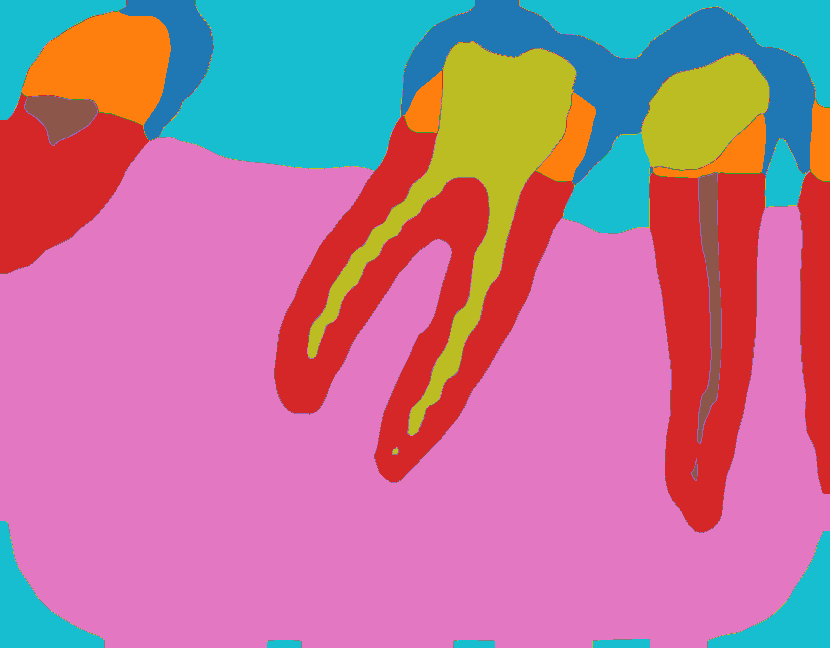}
    \end{subfigure}\hfill
    \begin{subfigure}{0.14\textwidth}
        \centering \includegraphics[width=\linewidth,trim={0.5cm 1.25cm 0.1cm 0},clip]{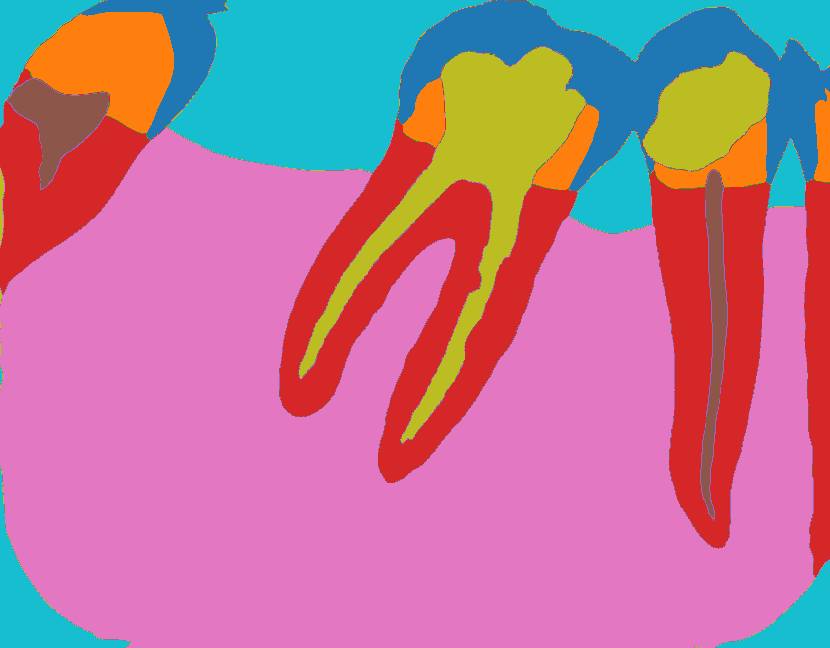}
    \end{subfigure}\hfill
    \begin{subfigure}{0.14\textwidth}
        \centering \includegraphics[width=\linewidth,trim={0.5cm 1.25cm 0.1cm 0},clip]{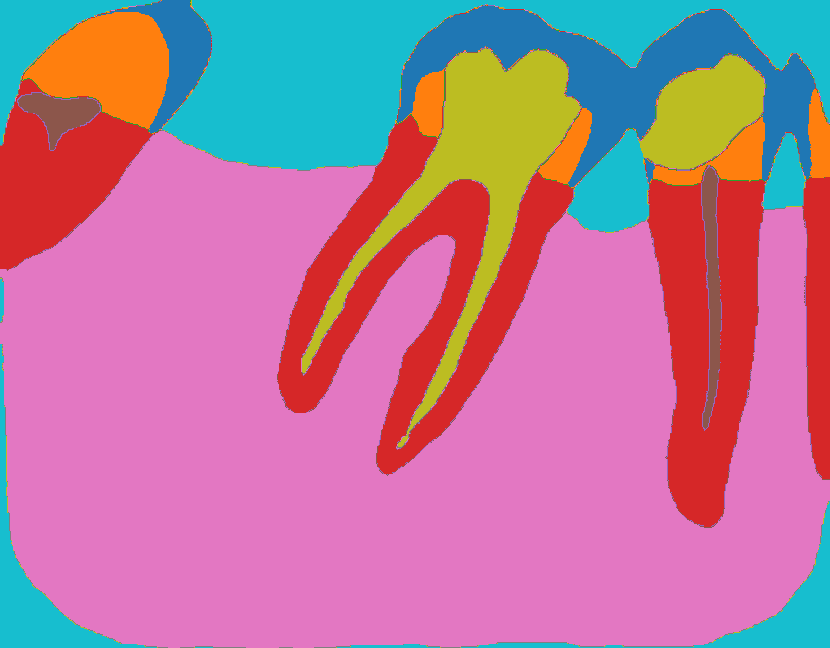}
    \end{subfigure}\hfill
    \begin{subfigure}{0.14\textwidth}
        \centering \includegraphics[width=\linewidth,trim={0.5cm 1.25cm 0.1cm 0},clip]{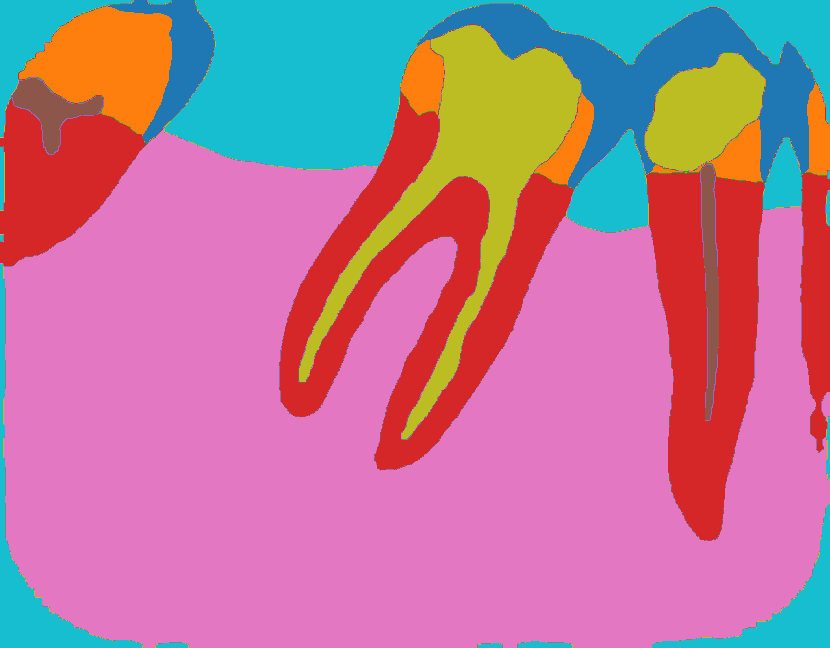}
    \end{subfigure}

    \vspace{0.05em}

    \begin{subfigure}{0.14\textwidth}
        \centering \includegraphics[width=\linewidth,trim={0.1cm 0.5cm 0.1cm 0.7cm},clip]{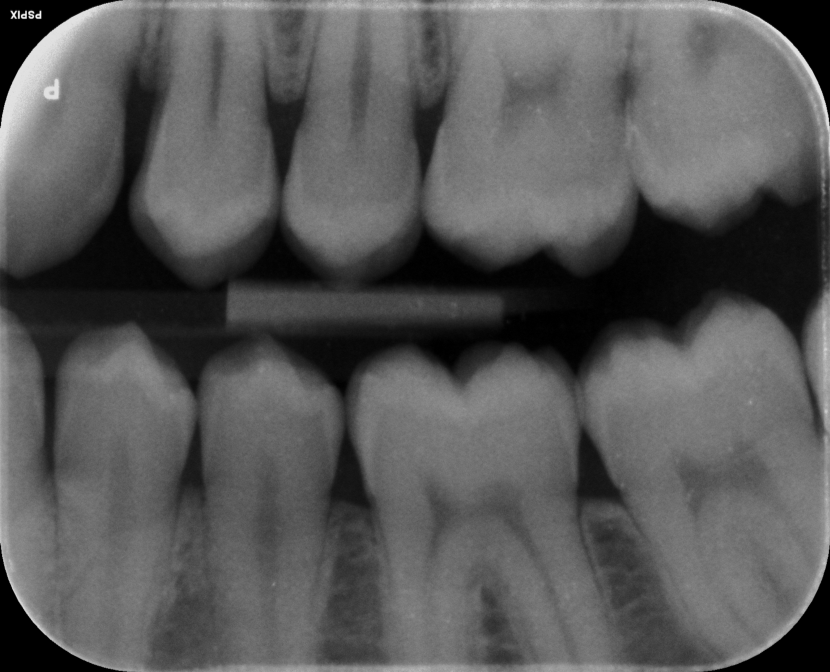}
        \caption{Input}
    \end{subfigure}\hfill
    \begin{subfigure}{0.14\textwidth}
        \centering \includegraphics[width=\linewidth,trim={0.5cm 2cm 0.3cm 3cm},clip]{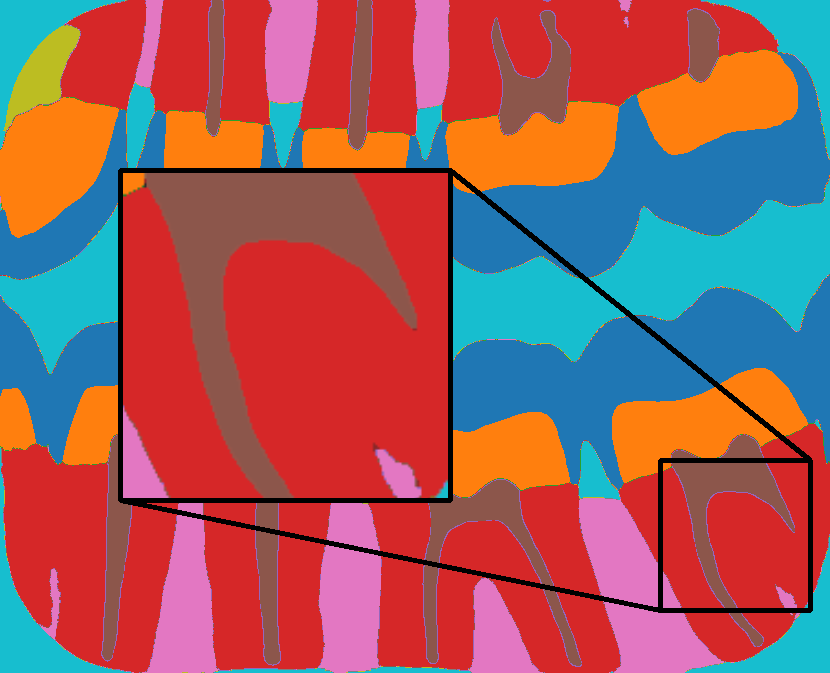}
        \caption{UNet}
    \end{subfigure}\hfill
    \begin{subfigure}{0.14\textwidth}
        \centering \includegraphics[width=\linewidth,trim={0.5cm 2cm 0.3cm 3cm},clip]{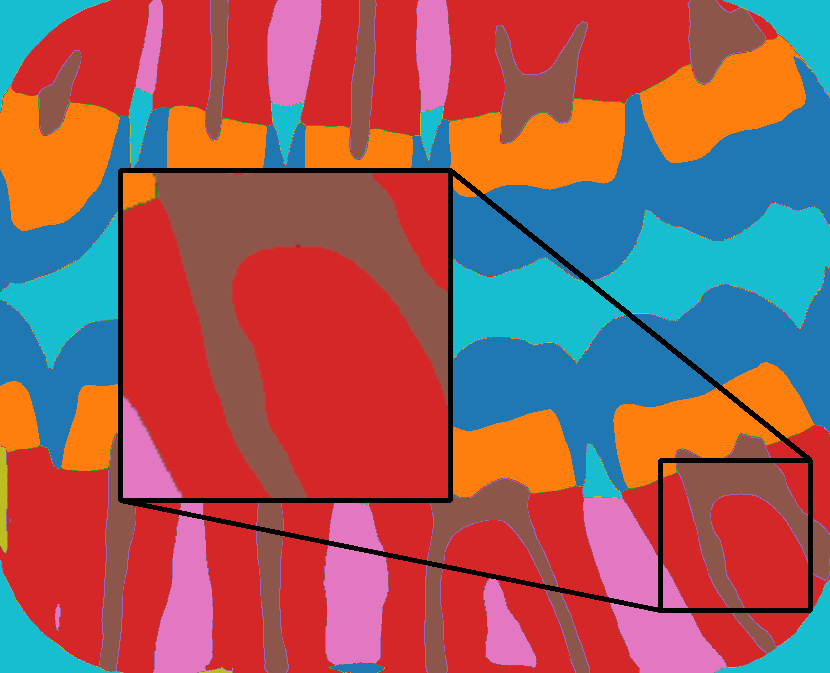}
        \caption{SETR}
    \end{subfigure}\hfill
    \begin{subfigure}{0.14\textwidth}
        \centering \includegraphics[width=\linewidth,trim={0.5cm 2cm 0.3cm 3cm},clip]{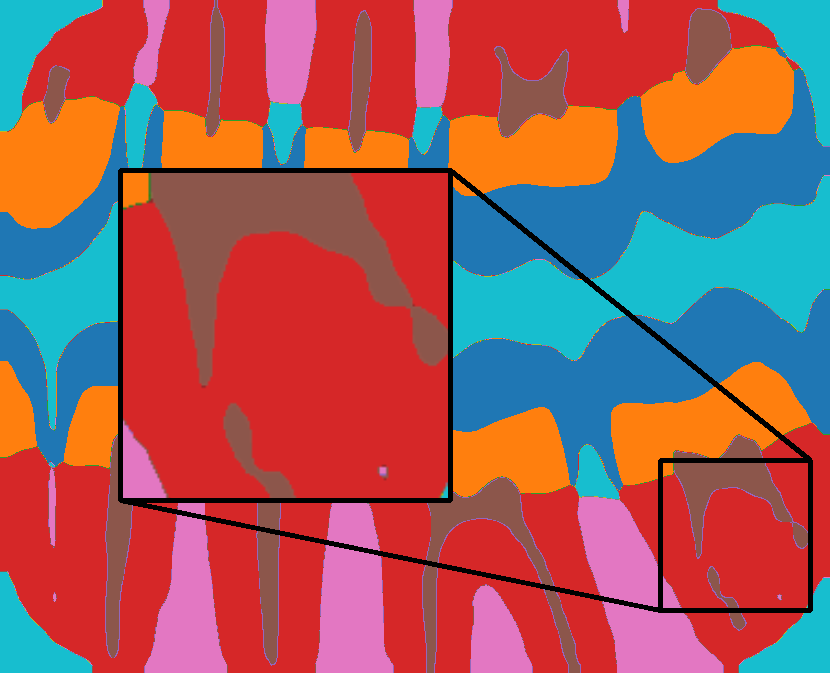}
        \caption{PIDNet}
    \end{subfigure}\hfill
    \begin{subfigure}{0.14\textwidth}
        \centering \includegraphics[width=\linewidth,trim={0.5cm 2cm 0.3cm 3cm},clip]{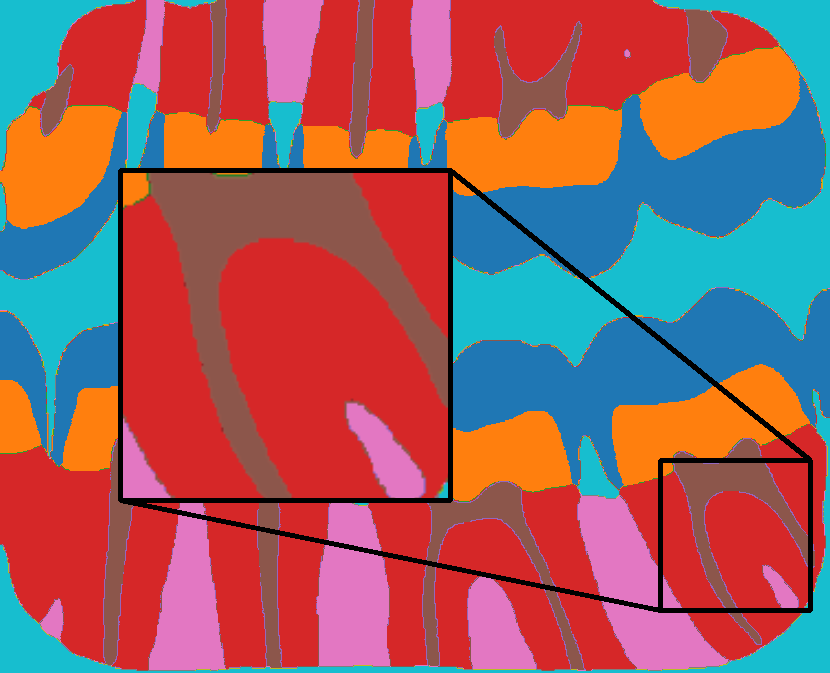}
        \caption{DentalX}
    \end{subfigure}\hfill
    \begin{subfigure}{0.14\textwidth}
        \centering \includegraphics[width=\linewidth,trim={0.5cm 2cm 0.3cm 3cm},clip]{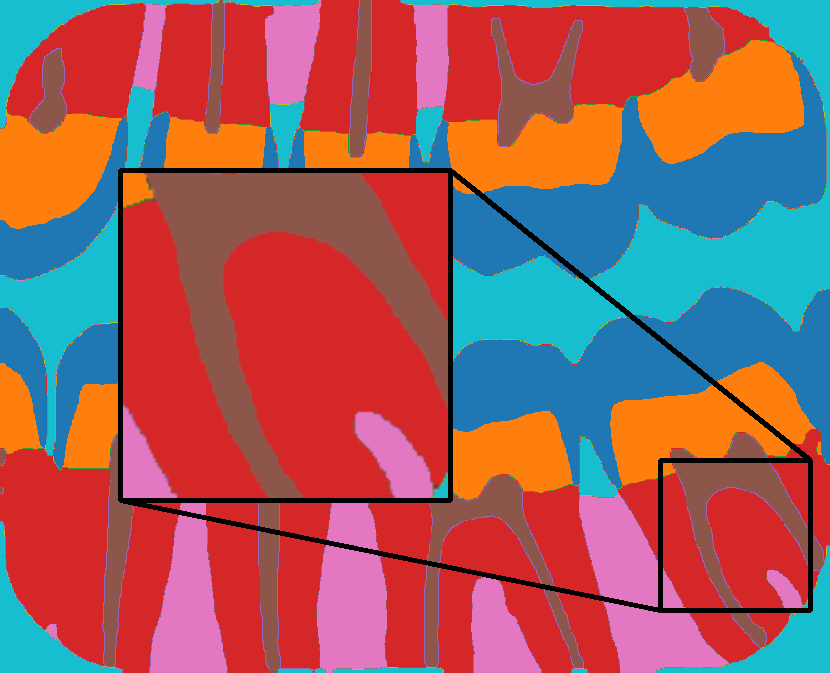}
        \caption{GT}
    \end{subfigure}
    \caption{(a) Input image. (b) - (e) Dental anatomy segmentation results. (f) Ground truth (GT) segmentation mask.}
    \label{fig:seg_qualitative}
\end{figure*}

\noindent\textbf{Comparative Results:}
As shown in \Cref{tab:result_dd}, DentalX achieves 45.9 AP$^{.5}$ and 21.0 AP$^{.5:.95}$, outperforming other baseline detectors in disease detection. 
It surpasses YOLOX by 5.2 AP$^{.5}$, 
while incurring only a slight drop in the inference frame per second (FPS) (6\% less than YOLOX). Due to the large number of parameters, DINOv3 fails to generalize well to our small dental dataset, achieving only 30.1 AP$^{.5}$.
Moreover, our method significantly outperforms the Faster R-CNN model, a commonly used object detector for disease detection in previous studies \cite{Chen2021caries_bl,zhu2022caries,Lee2022caries_pano,Liu2022marginalbl,ba2023perilesion_pano} by 11.7 AP$^{.5}$. \cref{fig:det_qualitative} visualizes the disease detection result of the object detectors. DentalX is consistently accurate with fewer false positives, further validating its capability to constrain the detected dental disease based on the structural context.

Besides, the performance of the proposed method in dental anatomy segmentation is summarized in \Cref{tab:result_as}. DentalX achieves mIoU, mDice, and mAcc of 86.3, 92.5, and 94.2, respectively, outperforming other baselines. This is because the segmentation task benefits from the better-learned features of the shared FPN backbone, leading to increased segmentation performance (See Ablation Study).
Moreover, as shown in \cref{fig:seg_qualitative}, DentalX produces anatomy masks that are more pixel-accurate than other baselines.

To quantify the utility of incorporating dental domain knowledge in the inference phase, we choose the best-performing models of each task (\ie YOLOX and SETR) and manually remove detected dental diseases that do not conform to the domain rules.
In particular, we specify the dental anatomies where each type of dental disease cannot occur, then remove false detections of YOLOX based on the segmentation output of SETR with a 0.5 IoU threshold. The post-inference method yields a slightly worse AP$^{.5}$ of 39.2,  demonstrating that incorporating domain knowledge post-inference is inferior to DentalX, which integrates the domain knowledge and structural context during model training.

\noindent\textbf{Ablation Study:}
\Cref{tab:ablation} presents the quantitative results of the ablation experiments. 
`Det. only' and `Seg. only' indicate the single-task settings of the dental disease detection and anatomy segmentation tasks, respectively, while the joint learning setting with and without extracting the structural context is represented as `Det. \& Seg.' in \Cref{tab:ablation}. We observe that under the joint learning setting, the performance of both tasks improves significantly compared to the single-task settings. 
Furthermore, extracting and utilizing the structural context improves the detection AP$^{.5}$ metric from 43.3 to 45.9 and the mIoU metric from 85.5 to 86.3. This validates that simply learning both tasks yields more representative backbone features, and the SCE module is effective in supporting the detection branch with anatomy context.

\begin{table}[tb]
    \caption{Ablation study. The `Struct. Context' column indicates the extraction of structural context.
    }
    \label{tab:ablation}
    \centering \setlength{\tabcolsep}{4pt}%
    \resizebox{\linewidth}{!}{
    \begin{tabular}{l|c|ccc|ccc}
         \hline
         Setting & \makecell[c]{Struct.\\ Context} & AP$^{.5}$ & AP$^{.75}$ & AP$^{.5:.95}$ & mIoU & mDice & mAcc \\
         \hline \hline
         Det. only & - & 40.7 & 13.1 & 18.1 & - & - & - \\
         \hline
         Seg. only & - & - & - & - & 80.1 & 91.7 & 88.5 \\
         \hline
         \multirow{2}{*}{Det. \& Seg.} & \crossmark & 43.3 & 15.0 & 19.4 & 85.5 & \textbf{93.6} & 92.1 \\
          & \checkmark & \textbf{45.9} & \textbf{17.1} & \textbf{21.0} & \textbf{86.3} & 92.5 & \textbf{94.2} \\
         \hline
    \end{tabular}
    }
\end{table}

%% file: sec/4_conclusion.tex
\section{Conclusion}
In this paper, we propose DentalX, a novel context-aware dental disease detection model that utilizes oral structure information to mitigate inherent visual ambiguity in radiograph images and improve the detection of subtle dental diseases. We introduce a structural context extraction module that learns dental anatomy segmentation as an auxiliary task to extract rich structural context.
DentalX then incorporates the extracted anatomy context and features into the primary dental disease detection task, allowing the model to capture the relationship between dental anatomy and dental disease. This is not achievable with prior works, as they tackle dental disease detection and anatomy segmentation tasks separately. 
Experimental results demonstrate that DentalX outperforms previous approaches, highlighting the mutual improvement by jointly learning both tasks and integrating structural context in dental disease detection.